\documentclass[journal]{IEEEtran}
\usepackage{microtype}
\usepackage{graphicx}
\usepackage{subfigure}
\usepackage{booktabs} 
\usepackage[pagebackref=false,breaklinks=true,letterpaper=true,bookmarks=false,citecolor={black}, colorlinks=true,linkcolor={black}]{hyperref}
\usepackage{epsfig}
\usepackage{graphicx}
\usepackage{amsmath}
\usepackage{amssymb}
\usepackage{color}
\usepackage[normalem]{ulem}
\usepackage{calligra}
\usepackage{multirow}
\usepackage{verbatim}
\usepackage[linesnumbered,ruled]{algorithm2e}
\usepackage{bbm}

\def\0{{\bf 0}}
\def\1{{\bf 1}}

\def\mbR{{\mathbb R}}

\definecolor{mygray}{gray}{.9}
\usepackage{amsmath} 
\usepackage{amssymb}  
\usepackage[table ]{ xcolor}

\begin{document}
\title{Fine-grained Classification via Categorical Memory Networks} 

\author{Weijian~Deng,
        Joshua Marsh,
        Stephen Gould,
  and Liang Zheng 
\IEEEcompsocitemizethanks{
\IEEEcompsocthanksitem W. Deng, J. Marsh, S. Gould, and L. Zheng (corresponding author) are with Australian National University, CBR, Australia.
\IEEEcompsocthanksitem Email: {firstname.lastname}@anu.edu.au
}}

\IEEEtitleabstractindextext{%
\begin{abstract}
Motivated by the desire to exploit patterns shared across classes, we present a simple yet effective class-specific memory module for fine-grained feature learning.
The memory module stores the prototypical feature representation for each category as a moving average. 
We hypothesize that the combination of similarities with respect to each category is itself a useful discriminative cue. 
To detect these similarities, we use attention as a querying mechanism. 
The attention scores with respect to each class prototype are used as weights to combine prototypes via weighted sum, producing a uniquely tailored response feature representation for a given input. 
The original and response features are combined to produce an augmented feature for classification.
We integrate our class-specific memory module into a standard convolutional neural network, yielding a Categorical Memory Network.
Our memory module significantly improves accuracy over baseline CNNs, achieving competitive accuracy with state-of-the-art methods on four benchmarks,  including CUB-200-2011, Stanford Cars, FGVC Aircraft, and NABirds.

\end{abstract}

}

\maketitle

\IEEEdisplaynontitleabstractindextext
\IEEEpeerreviewmaketitle

\section{Introduction}\label{sec:introduction}
\IEEEPARstart{T}{his} article studies fine-grained classification, one of the most inherently challenging visual recognition tasks. 
Standard classification techniques are able to rely upon large-scale global features to differentiate classes. Meanwhile, fine-grained datasets simultaneously exhibit both subtle inter-class variation, such as global structure and visual appearance, and large intra-class variation, such as pose. This has led most methods to rely on \emph{local differences} to provide discriminative cues \cite{yang2018learning,zheng2017learning,fu2017look,recasens2018learning}. We propose an orthogonal research direction that instead seeks to exploit \emph{inter-class similarities}, which is an essential characteristic of the fine-grained classification.

\begin{figure}[t]
\begin{center}
\includegraphics[width=1\linewidth]{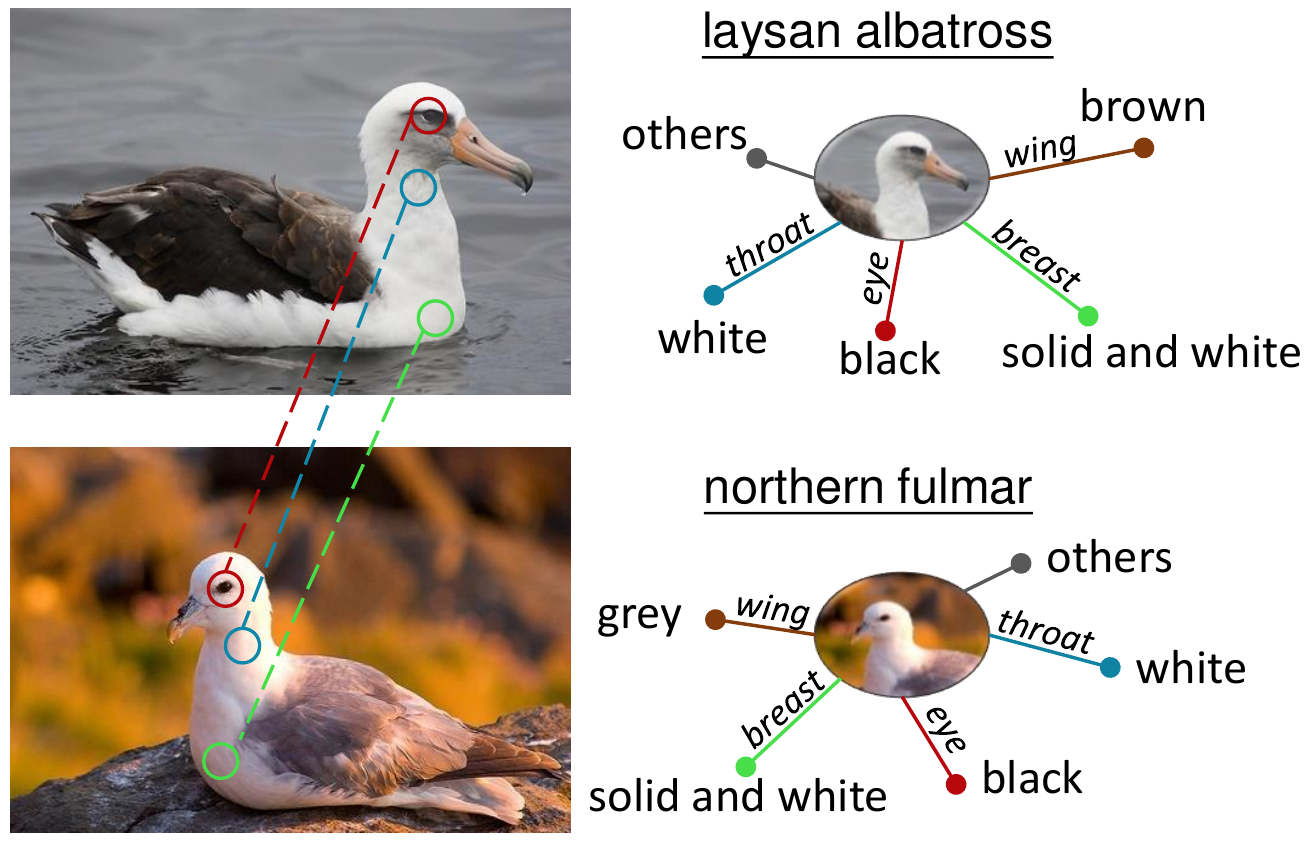}
\end{center}
\caption{Visual example of the correlation between fine-grained classes. Though the two birds belong to different subspecies, they are semantically and visually similar. Both share some common attributes (the first column), \emph{e.g.}, eye color, breast pattern, and throat color. This similarity indicates their representations will share some underlying semantic cues.}
\label{fig:motivation}
\end{figure}

Convolutional neural networks (CNNs) tend to learn features that discard some semantic information in favour of more discriminative visual patterns. In the case of fine-grained classification where discriminative cues are particularly subtle, the loss of semantic information may lead to a lack of generalization. Our {insight} is that encouraging the network to represent similarities in addition to unique discriminative cues is beneficial for robust fine-grained feature learning. Consider the case study of bird classification. Different species share both semantic and visual attributes such as eye color, throat color, and breast pattern, as illustrated in Figure \ref{fig:motivation}. 
Instead of discarding these shared semantics, we pose they could be used to produce more informative and robust features.
A natural question arises: how can we use these shared inter-class semantics to improve the fine-grained classification? 

%

%
Motivated by this, we propose a class-specific memory module that gives a network direct access to the statistics of class-specific feature representations. Our module is interpretable, with each slot in memory corresponding to the average feature of a given category, which we refer to as a class \emph{prototype}. The class-specific memory module can be directly integrated into popular CNN architectures without any additional modifications. It is trained in an end-to-end supervised manner, with the memory module updated while training along with the rest of the network parameters. Given a labelled feature, we update the corresponding memory slot via moving average. Thus, each memory slot is encouraged to represent the prototypical information of its corresponding class.

Intuitively, each class should exhibit a specific distribution of similarities with respect to the other classes. We hypothesize that this distribution is itself a useful identifier of class.
The rich semantic correlation between classes provides beneficial guidance for fine-grained feature learning. Specifically, we want to give neural networks the ability to perceive the shared semantics across classes while preserving the unique and subtle cues of the given input sample. To this end, we propose an attention-based method of addressing memory, where each prototype is retrieved based on its relevance to the current input feature. We define relevance as the attention score based on cosine similarity in the learned feature space. We use the attention score to retrieve relevant class prototypes with respect to a given input, producing a response feature. Finally, the input feature and the response feature are combined, resulting in an augmented feature. This augmented feature is characterized by 1) expressing inter-class similarities of the input while 2) maintaining the individuality of input representations. 

The proposed method is simple, effective, adds no additional parameters, and can be applied across non-rigid (e.g, birds) and rigid (e.g, cars) visual datasets. Furthermore, the entire memory module, in practice a matrix, is itself negligible in terms of computational overhead (including reading/writing) and adds no additional trainable parameters.

The key contributions of this paper include:
\begin{itemize}
\item We introduce a novel Categorical Memory Network (CMN) for fine-grain visual classification. Our method incorporates both feature learning and inter-class knowledge to yield more expressive features. 
\item To leverage the knowledge of all categories, we introduce an attention-based memory addressing method to adaptively fuse prototypes.
\item CMN yields significant improvement over baselines, achieving competitive accuracy with the state-of-the-art accuracy on four fine-grained visual classification benchmarks.

\end{itemize} 

\section{Related Work}
\textbf{Fine-grained Feature Learning.} Fine-grained visual categorization is a special case of image classification wherein there exists significantly more inter-class similarities than standard classification tasks (e.g ImageNet). Fine-grained datasets share the internal opposing dynamics of subtle inter-class and large intra-class variation, which together represent the central challenge for this task. 

Deep Convolutional Neural Networks (CNNs) \cite{SimonyanZ14a,he2016deep,szegedy2015going} are unable to sufficiently represent discriminative information required to classify fine-grained images. To obtain more expressive representations, several feature encoding methods have been proposed \cite{lin2015bilinear,gao2016compact,kong2017low,yu2018hierarchical,luo2019cross,chang2020devil}. Luo \emph{et al.} \cite{luo2019cross} introduce a cross-layer regularizer to use multi-scale features for classification. Lin \emph{et al.} \cite{lin2015bilinear} present a bi-linear pooling method to extract features with two Convolutional Networks. Gao \emph{et al.} \cite{gao2016compact} introduce a bi-linear pooling in a kernelized framework, improving the computational efficiency of bi-linear pooling.
Kong \emph{et al.} \cite{kong2017low} introduce a low-rank bi-linear classifier to decrease the computation time. 
Furthermore, Yu \emph{et al.} \cite{yu2018hierarchical} propose a hierarchical bi-linear pooling approach to fuse multi-layer features for fine-grained classification.

Another area of research focuses on finding discriminative regions in images. Early studies \cite{zhang2014part,lin2015deep,huang2016part,zhang2016spda,goring2014nonparametric,he2019and} make use of bounding boxes and part annotations to localize discriminative regions. While such supervised annotations have proven beneficial to fine-grained feature learning, they are costly to obtain.
More recently, weakly supervised methods have emerged as a promising research area \cite{peng2017object,zhang2016weakly,yang2018learning,zheng2017learning,fu2017look,sermanet2014attention,sun2018multi,recasens2018learning,wang2018learning}. These methods use image category labels to localize object parts and extract part features for classification.
Yang \emph{et al.} \cite{yang2018learning} use a navigator network to detect informative regions. Zheng \emph{et al.} \cite{zheng2017learning} group ConvNet channels to detect various local patterns. Similarly, Fu \emph{et al.} \cite{fu2017look} recursively localizes parts at multiple scales while Sermanet \emph{et al.} \cite{sermanet2014attention} uses an attention mechanism to select a series of regions. Sun \emph{et al.} \cite{sun2018multi} introduce a one-squeeze multi-excitation module to learn region features. 
Moreover, Recasens \emph{et al.} \cite{recasens2018learning} recently proposed the use of saliency maps to localize informative regions. 

In addition to the above approaches, some works focus on the small inter-class variations \cite{dubey2018pairwise,dubey2018maximum,zhou2016fine}. Zhou \emph{et al.} \cite{zhou2016fine} exploit the rich relationships among categories through bipartite-graph labels. Dubey \emph{et al.} \cite{dubey2018pairwise} introduce a pairwise confusion loss to prevent CNNs from over-fitting. 
The confusion loss attempts to bring class conditional probability distributions closer to each other. This enables the network to preserve some shared semantic among classes. 
Our CMN explicitly encourage the network to leverage shared prototypical information across classes for enhancing the input features. The class-specific memory module gives a network direct access to the statistics of class-specific feature representations. Thus, the network can adaptively retrieve relevant feature prototypes of different classes to augment input features. Moreover, our method focuses on feature combinations after obtaining image features, and as such is compatible with the existing approaches.

\begin{figure*}[t]
\begin{center}
\includegraphics[width=1.0\linewidth]{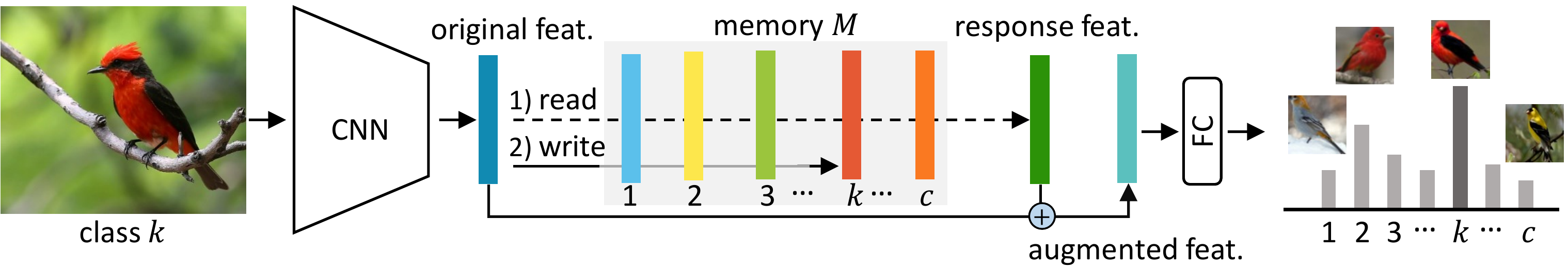}
\end{center}
\caption{Architecture of Categorical Memory Network (CMN) for fine-grained classification. CMN integrates a class-specific memory module into an existing convolution neural network architectures (\emph{e.g.}, ResNet-50). The module contains memory slots to record category prototypes. Given an original feature produced by the CNN as a query, we retrieve and combine the relevant prototypes, and then output a response feature. We then update the corresponding memory slot based on the label of the original feature. Finally, we augment original feature with response feature. The resulting augmented feature is used for classification.}
\label{fig:framework}
\end{figure*}

\textbf{Memory Networks.}
Memory networks \cite{WestonCB14,sukhbaatar2015end,graves2016hybrid} are a unique variant of neural networks that allow a network to explicitly read and write information to an external memory module. Inputs are sequentially saved to memory and retrieved based upon the relevance (usually calculated via attention) to the current network input.
Weston \emph{et al.} \cite{WestonCB14} introduced memory-augmented networks for the task of question answering. Santoro \emph{et al.} \cite{santoro2016meta} use an external memory to tackle meta-learning tasks. Li \emph{et al.} \cite{li2016learning} and Kim \emph{et al.} \cite{kim2018memorization} use memory networks for data generation.
The closely related concept of Neural Turing Machines \cite{graves2014neural} provides more complex functionality such as continuous memory representations and location and content-based access. 
We study memory networks in the context of fine-grained visual classification and introduce a categorical memory module to enhance the fine-grained feature learning. 

In the class-specific memory module, each row of the module corresponds to the prototypical information of a single class in the form of a moving average. This practice is relevant to prototypical networks \cite{hsu2018unsupervised,snell2017prototypical,wertheimer2019few} in few-shot setting. However, our work is significantly different from them. These works use  prototypes as class centers for the nearest-neighbor classifier. In comparison, we use relevant class prototypes to enhance the input features.

\section{Categorical Memory Network}
The proposed class-specific memory module can be directly integrated into an existing convolutional neural network without additional modifications. We refer to this network as a Categorical Memory Network (CMN). In practice, we insert the memory module after the last convolution layer of a convolutional neural network, \emph{e.g.}, ResNet-50 \cite{he2016deep}. 

An overview of CMN is shown in Figure \ref{fig:framework}.
We first pass the image through the convolutional layers and the final global average pooling, producing an original feature.
We then use this original feature as the input to the class-specific memory module to retrieve the relevant prototypes, leading to a response feature.
This response feature is then combined with the original feature via a weighted sum with learned coefficients, producing an inter-class similarity aware augmented feature.
Finally, this augmented feature is fed through a single fully connected layer and a softmax to obtain class predictions.

The network is trained in an end-to-end manner via class-based supervised learning, \emph{i.e.}, we use cross-entropy loss.
In the following section, we introduce the categorical memory module and detail its reading and writing procedures.

\subsection{Class-specific Memory Module}
Motivated by the intuition that giving a network access to leverage class-specific information is beneficial to fine-grain feature learning, we introduce a memory module to store the mean feature vector for each class. We refer to this as the feature prototype of a class. The memory module primarily consists of two operations, 1) writing the feature prototypes to memory module, and 2) computing response feature based on the current input feature. 

During \emph{training}, the memory slots, i.e rows, are read from based on the relevance to the given input. We then update memory slots based on the categories of given inputs in a mini-batch. 
During \emph{testing}, the memory is not updated and used only for reading.

\subsubsection{Memory Writing}
We implement memory as a matrix $\bf M\in \mbR^{C \times D}$, where C is the number of classes and D is the dimension of the prototype. Each row ${\bf m}_{i}$ of $\bf M$ represents the feature prototype of the $i$-th class.
Instead of updating the prototypes with respect to the entire training set, we update by moving average per mini-batch\footnote[1]{We perform the moving average update with the current mini-batch samples after memory reading.}.
Given a feature ${\bf f}_{i}$ from $i$-th category, the corresponding prototype ${\bf f}_{i}$ is computed as,  
\begin{equation}
   {\bf m}_{i} \leftarrow {\bf m}_{i} + \beta({\bf f}_{i}-{\bf m}_{i}),
    \label{moving}
\end{equation}
where $\beta \in (0, 1)$  is a coefficient that controls the update rate of prototype ${\bf m}_{i}$.
\begin{figure}[t]
\begin{center}
\includegraphics[width=1\linewidth]{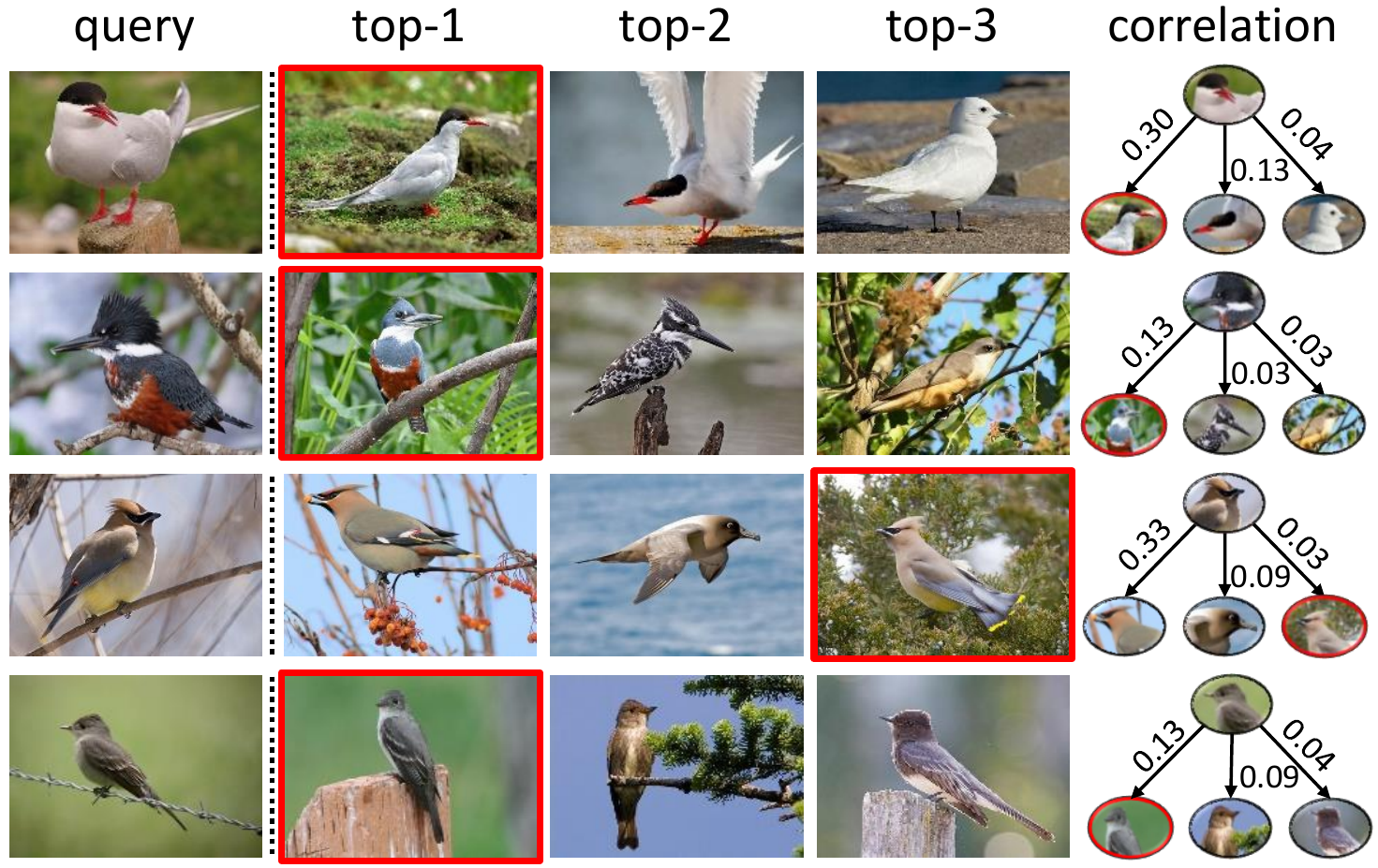}
\end{center}
\caption{Visualization of memory addressing. Our module is able to retrieve the relevant class prototypes for a given query image (the first column). In this figure, we show exemplar images from the three most relevant prototypes (the second to fourth columns) with respect to a given query image. We highlight red the prototype of the same class. In the last column, we show the attention scores between the input image and each retrieved prototype.}
\label{fig:3}
\end{figure}

\subsubsection{Memory Reading} \label{read}
Our goal is to borrow some prototypical knowledge from relevant classes to augment the input feature, yielding more informative and comprehensive representations.

We use attention as our addressing mechanism. Given an input feature as a query, our module produces a response feature that is a uniquely tailored linear combination of all class prototypes in memory.  
Specifically, we calculate the attention score of the input feature $\bf f$ with respect to each prototype ${\bf m}_{i}$ (for $i \in \{1, \ldots, C\}$). These attention scores are used as weights to combine the prototypes, resulting in a response feature.
We define the memory addressing process as, 
\begin{equation}
   \widetilde{\bf m} = \sum_{i=1}^{C}w_{i}{\bf m}_{i},
    \label{combination}
\end{equation}
where $w_{i}$ is the attention score for the prototype ${\bf m}_{i}$, and $\widetilde{\bf m}$ is the response feature.

\textbf{Attention Score.} Intuitively, if a prototype is close to input feature in the representation space, the prototype is more likely to contain relevant information and is weighted accordingly.
In practice, we use cosine similarity distance as our similarity metric. We normalize the similarity scores by a softmax function, giving the final attention scores. This process is defined as, 

\begin{equation}
    w_{i} =  \frac{\exp({\bf f}{\cdot}{\bf m}_{i} / \tau)}{\sum_{j=1}^{C}\exp({\bf f}{\cdot}{\bf m}_{j} / \tau)},
    \label{attention}
\end{equation}
where $w_{i}$ is the normalized attention score between input feature $\bf f$ and prototype ${\bf m}_{i}$ of class $i$, the temperature $\tau$ is empirically set as 0.1. In our case, each prototype represents one fine-grained class. Classes with strong similarities to an input image should receive higher attention scores and hence be weighted higher in the memory response.

An illustration of this is shown in Figure \ref{fig:3}. 
We show exemplary images from the three most relevant prototypes for a given input image query. 
The query image and the images from the classes represented by the retrieved prototypes share visual and semantic similarities, \emph{e.g.}, color, wing pattern, bill shape, and breast pattern. Moreover, we also observe that the input image is more likely to select the prototype corresponding to its class.
Based on this, we think that our attention-based memory addressing method is effective to detect relevant class prototypes for the given image.

\begin{figure}[t]
\begin{center}
\includegraphics[width=0.85\linewidth]{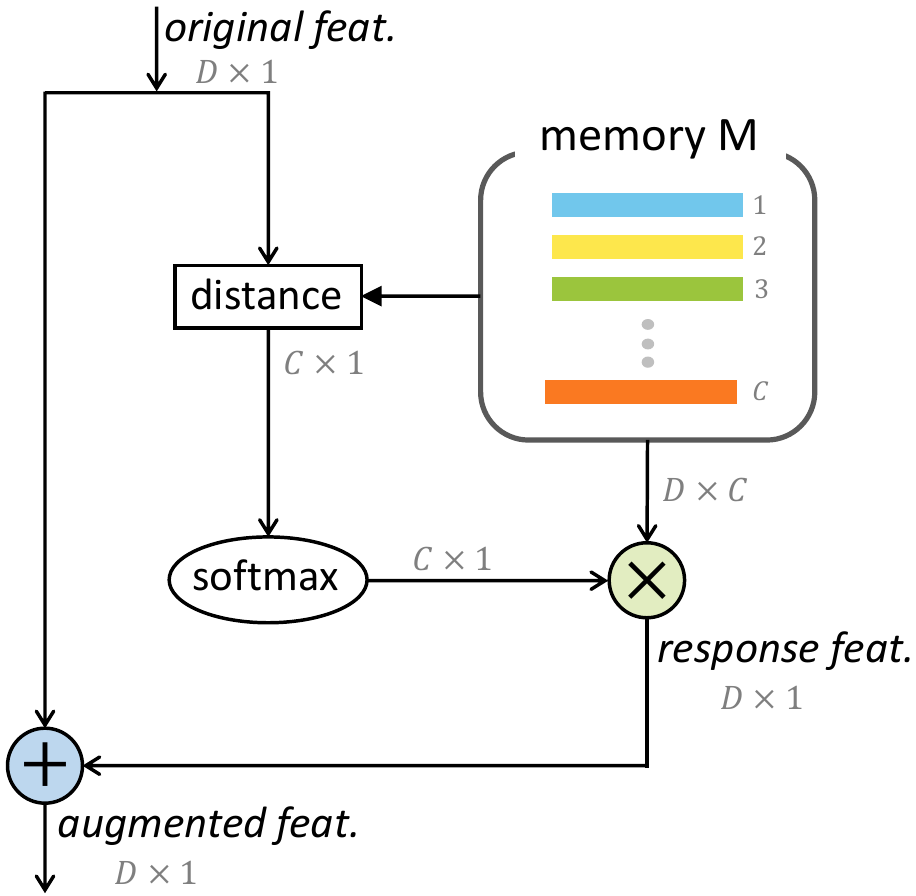}
\end{center}
\caption{Pipeline of memory reading process. Given an original feature, we calculate its Euclidean distance to each class prototype, which we then normalize to produce attention scores. After that, we use attention scores as weights to combine prototypes, resulting a response feature (response feat.). Finally, we combine the original feature and the response feature by using a weighted summation, yielding an augmented feature (augmented feat.) that is used for classification. In this figure, $\otimes$ is matrix multiplication, and $\oplus$ is broadcast element-wise addition.}
\label{fig:4}
\end{figure}

 \textbf{Feature Augmentation.} The original feature, which is instance-specific, and response feature, which is class-specific, are combined to produce an augmented feature representation that is used for classification.
 Specifically, we use the following summation:
\begin{equation}
    {\bf f}_{aug} =  \bf f +  \widetilde{\bf m}.
    \label{final feature}
\end{equation} We also attempted more sophisticated methods to combine two features such as element-wise scaling and fully-connected layers, but these did not improve performance. An example of memory reading pipeline is illustrated in Figure \ref{fig:4}. The reading process can be simply done by matrix operations.

\subsection{Discussion}
\textbf{Working mechanism.} The learning mechanism of our method is that it makes the network adaptively exploit the knowledge of each class to generate more expressive representations. The advantage of the mechanism is two-fold, 1) it encourages neural networks to preserve shared semantic information, enabling a comprehensive understanding of fine-grained classes; 2) it allows networks to utilize external knowledge from relevant classes to better represent a given image, resulting in more robust and informative representation. The result is a final feature that captures both instance specific characteristics and the inter-class semantic relationships for the given input. Specifically, for the most indistinguishable classes, CMN will learn to focus more on instance specific characteristics. Thus, the subtle differences essential for classifying these classes can be better represented. 

It is worth noting that our method does not require any side information such as structural labels \cite{xie2015hyper,wang2015multiple} or shared attributes \cite{zhou2016fine}.
Instead, our network can learn to adaptively select shared cues across all categories for a given input. Moreover, our method is compatible with several part-localization based methods such as  \cite{zheng2019looking,recasens2018learning}, and hence has the potential to further improve recognition. 

\textbf{Interpretability.} The physical meaning of the proposed class-specific memory module is very clear. The module is used to store and share representative characteristics between classes. Specifically, each row of the module corresponds to the prototypical information of a single class in the form of a moving average.
In addition, the attention-based memory reading is based on the relevance between given inputs and prototypes and hence is also interpretable. 
Note that, in few-shot setting, class prototypes are used as class centers for constructing the nearest-neighbor classifier \cite{hsu2018unsupervised,snell2017prototypical,wertheimer2019few}. Our work is significantly different from them: we use relevant class prototypes to enhance the input features.

\begin{figure}[t]
\begin{center}
\includegraphics[width=1\linewidth]{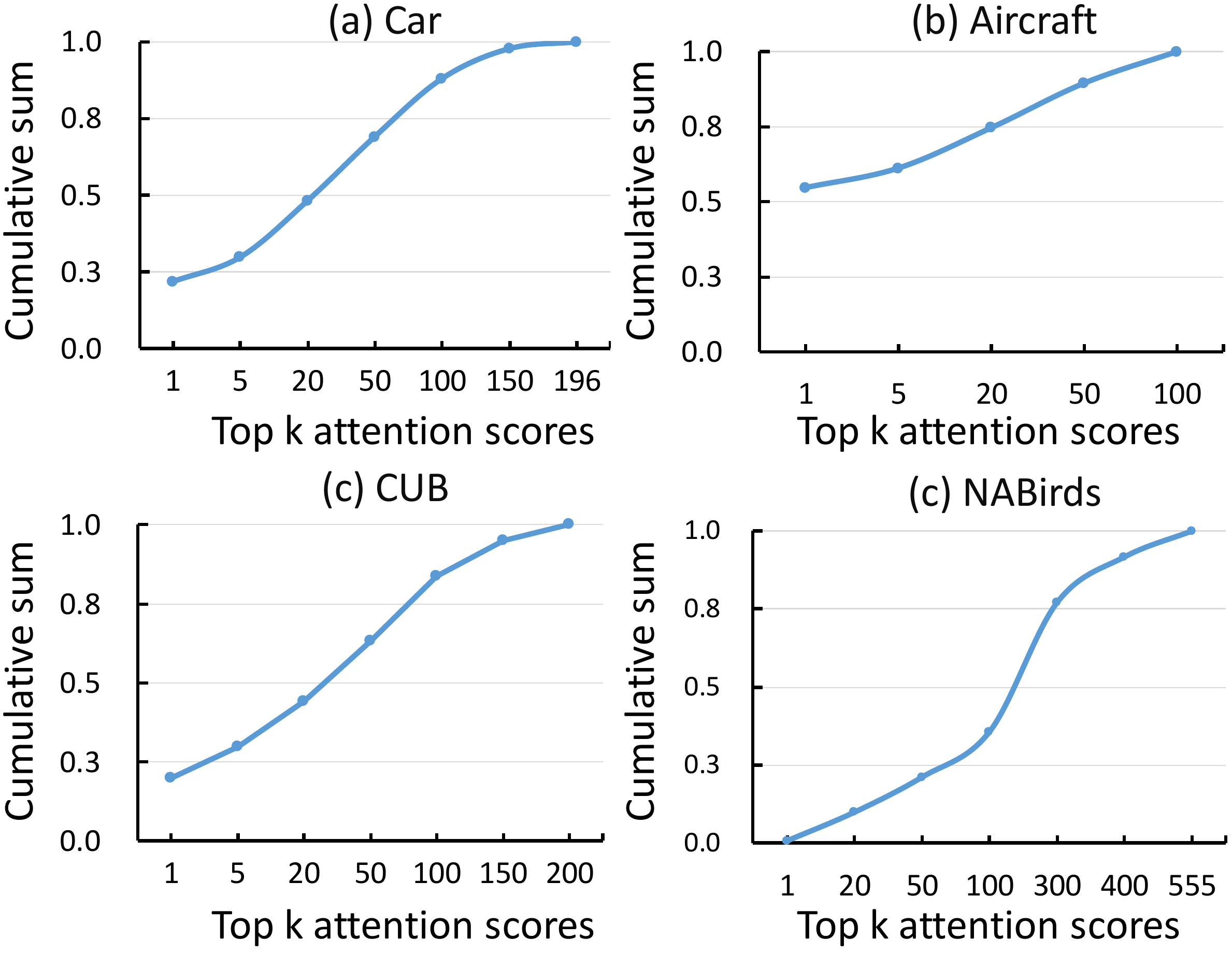}
\end{center}
\caption{Cumulative sum of attention scores for top-k highest matching prototypes in memory averaged over all samples in the (a) Car, (b) CUB, (c) Aircraft, and (d) NABirds datasets. Notice that on the right end of each graph, the attention scores of all classes sum to one.}
\label{fig7}
\end{figure}

\textbf{Attention distribution.} We visualize the cumulative sum of attention scores for top-k most similar prototypes in memory averaged over all samples in a dataset. The cumulative sum curve directly represents the distribution of attention scores on a specific dataset. We show the cumulative sum curves of four datasets in Figure \ref{fig7}.
We observe that approximately 80\% of the response feature is made up of only half the prototypes. This is consistent with our intuition that more relevant prototypes contribute more information to augment the input feature. 

\textbf{Efficiency.} Our method introduces a memory module in the form of a $C \times D$ matrix. It involves an additional matrix multiplication, which is computationally negligible. Thus, it has very similar inference time with baseline even when the number of classes $C$ is large.
CMN requires only a single forward pass, while the most localization-based methods \cite{yang2018learning,zheng2019looking} require multiple forward passes. Comparing with them, the computational cost of CMN is much less.

\setlength{\tabcolsep}{4pt}
\begin{table}[t]
\caption{Statistics of three fine-grained
classification datasets.}
\small
    \begin{center}
    \begin{tabular}{|c|c|c|c|}
    \hline
    Dataset & $\#$Train Set & $\#$Test Set& $\#$Category  \\
    \hline
    \hline
    FGVC Aircraft \cite{maji2013fine} & $6,667$ & $3,333$ & $100$ \\
    \hline
    CUB-200-2011 \cite{wah2011caltech} & $5,994$ & $5,794$ & $200$ \\
    \hline
    Stanford Cars \cite{krause20133d}  & $8,144$ & $8,041$ & $196$ \\
    \hline
    NABirds \cite{van2015building} & 23,929 & 24,633 & 555 \\
    \hline
    \end{tabular}
    \end{center}
\label{data_stat}
\end{table}
\section{Experiment}
\subsection{Datasets}
We comprehensively evaluate our method on four datasets, including CUB-200-2011 \cite{wah2011caltech}, Stanford Cars \cite{krause20133d}, FGVC Aircraft \cite{maji2013fine}, and NABirds \cite{van2015building}, respectively. 

\textbf{CUB-200-2011} contains 200 birds categories
with roughly 30 training images per category. It has 5994 training images and 5794 testing images.
\textbf{FGVC-Aircraft} is comprised of 10000 images over 100 categories, with a ratio of the training set to testing set of roughly $2:1$. 
\textbf{Stanford Car} is made up of 196 categories of cars, with 8144 examples in training set and 8041 examples in the testing set.
\textbf{NABirds} is a large scale dataset, consisting of 555 classes. It is made up of $23,929$ training images and $24,633$ testing images.

In the experiment, we follow the same train/test splits in Table \ref{data_stat}. Our method does not utilize any auxiliary supervision methods such as extra annotations and relies solely on image labels for the learning supervision. We report the top-1 accuracy in all experiments.

\subsection{Implementation Details}
\textbf{Implementation Details.}
We implement our method in Pytorch \cite{paszke2017automatic} and train all models on a single Tesla P-100 GPU. We adopt ResNet-50 \cite{he2016deep} pre-trained on ImageNet \cite{deng2009imagenet} as our backbone network.
The proposed method is trained using the SGD optimizer with momentum 0.9, weight decay 1e-4 and a mini-batch size of 16. We use a learning rate 4e-3 for the backbone and $5\times$ multiplier for the newly added layer. 
We train the network for 60 epochs and decay the learning rate by 0.1 after 40 epochs.
We preprocess images to $448 \times 448$ and use random horizontal flipping with a probability of $0.5$ for data augmentation. Moreover, we set coefficient $\beta$ in Eq. \ref{moving} as 0.9. 
The entire network is trained in an end-to-end manner via the supervision of image labels. The network \emph{does not} require any special initialization, multiple training stages, and part or bounding box annotations. We report the top-1 classification accuracy from the last epoch.

\setlength{\tabcolsep}{8pt}
\begin{table}[t]
\caption{Comparison of various methods in terms of top-1 accuracy (\%) on \emph{rigid} \textbf{FGVC-Aircraft}. * denotes our re-implementation.}
\small
\begin{center}
\begin{tabular}{|l|c|c|c|c|}
\hline
\multicolumn{1}{|l|}{\multirow{2}{*}{Methods}}& \multicolumn{1}{l|}{\multirow{2}{*}{Backbone}}& \multicolumn{1}{l|}{\multirow{2}{*}{Accuracy (\%)}}\\
\multicolumn{1}{|c|}{} & {}  &\\
\hline\hline
BilinearCNN \cite{LinRM18} & VGG-16 & 84.1  \\
Low-rank B-CNN \cite{kong2017low}& VGG-16& 87.3 \\
Kernel-Pooling \cite{cui2017kernel}&VGG-16& 86.9  \\
DFL-CNN \cite{wang2018learning}&VGG-16&91.1 \\
\hline
MA-CNN \cite{zheng2017learning}& VGG-19& 89.9 \\
\hline
Kernel-Pooling \cite{cui2017kernel}&ResNet-50& 85.7 \\
FT-ResNet \cite{he2016deep}* & ResNet-50 &90.0 \\
DFL-CNN \cite{wang2018learning}& ResNet-50& 91.7 \\
NTS-Net \cite{yang2018learning} & ResNet-50& 91.4 \\
DCL \cite{Chen_2019_CVPR} & ResNet-50 &93.0   \\
Cross-X \cite{luo2019cross} & ResNet-50 & 92.6 \\
LIO \cite{Zhou_2020_CVPR} & ResNet-50 &92.7 \\
ACNet \cite{ji2020attention} & ResNet-50 &92.4 \\
\hline
\rowcolor{mygray}{CMN (ours)} & ResNet-50  &  {93.8} \\
\hline
\end{tabular}
\end{center}
\label{tab2}
\end{table}

\subsection{Performance Comparison}
\textbf{Compared methods.} We compare the proposed method with ResNet-50 based methods and also include the best results of VGG based methods. For a fair comparison, we do not include approaches using additional data or annotations. The baseline is fine-tuning from ResNet-50 (FT-ResNet) \cite{he2016deep}, which is used in NTS-Net \cite{yang2018learning} and TASN \cite{zheng2019looking}.

\textbf{Comparison on FGVC-Aircraft.} In Table \ref{tab2}, we compare our method (CMN) with the existing methods on Aircraft. CMN achieves 93.8\% accuracy, which is $+3.8\%$ higher over baseline (FT-ResNet). The achieved accuracy is highest on the Aircraft. This indicates that the proposed class-specific memory module is beneficial for improving feature representations. Moreover, CMN outperforms LIO \cite{Zhou_2020_CVPR} and DCL \cite{Chen_2019_CVPR} by 1.1\% and 0.8\%, respectively. In addition, our method surpasses ACNet \cite{ji2020attention} by 1.4\%. Compared with TASN \cite{zheng2019looking} and NTS-Net \cite{yang2018learning}, CMN also gains competitive accuracy, showing its effectiveness for Aircraft classification.

\textbf{Comparison on Stanford Cars.} As shown in Table \ref{tab3}, CMN gains 2.4\% improvement over baseline (FT-ResNet). Compared with the state-of-the-art methods, CMN can achieve competitive accuracy. For example, it is 0.3\% and 0.4\% higher than ACNet \cite{ji2020attention} and DCL \cite{wang2018learning}, respectively. 
Moreover, our methods is also higher than other VGG based approaches, such as MA-CNN \cite{zheng2017learning} and Kernel-Pooling \cite{cui2017kernel}.
The above comparisons demonstrate our method is effective for the two rigid datasets with a significant structural variation. It is worthy of noting that CMN \emph{does not} require multiple forward passes like the most localization-based methods \cite{yang2018learning,zheng2019looking}. Therefore, we think the proposed method is efficient and effective for fine-grained feature learning.

\textbf{Comparison on CUB-200-2011.} We additionally report the results on non-rigid CUB-200-2011 shown in Table \ref{tab3}. We show that our CMN achieves 2.4\% improvement over baseline (FT-ResNet) and arrives at 88.2\%. This result is also comparable with the current state-of-the-art method ACNet \cite{ji2020attention} (88.2\% vs. 88.1\%). It is worth noting that CMN outperforms ACNet by $+1.4\%$ and $+0.3\%$ on Aircraft and Car, respectively.
Compared with TASN \cite{zheng2019looking}, our method is also competitive (88.2\% vs. 87.9\%). In addition, TASN and NTS-Net \cite{yang2018learning} require multiple forward passes to extract region features, making them significantly more computationally prohibitive. CMN requires only a single forward pass through the network, and thus is relatively efficient. Moreover, we think CMN would be compatible with these part-localization based methods, and further achieve improvements.
Our method also gains comparable accuracy with Cross-X \cite{luo2019cross}. Cross-X introduces a cross-layer regularizer to leverage multi-scale features from different layers, which we believe if combined with our method could also result in an improvement in classification accuracy.

\setlength{\tabcolsep}{6pt}
\begin{table}[t]
\caption{Comparison of various methods in terms of top-1 accuracy (\%) on \emph{rigid} \textbf{Stanford Cars}. * denotes our re-implementation.}
\small
\begin{center}
\begin{tabular}{|l|c|c|c|c|}
\hline
\multicolumn{1}{|l|}{\multirow{2}{*}{Methods}}& \multicolumn{1}{l|}{\multirow{2}{*}{Backbone}}& \multicolumn{1}{l|}{\multirow{2}{*}{Accuracy (\%)}}\\
\multicolumn{1}{|c|}{} & {}  &\\
\hline\hline
BilinearCNN \cite{LinRM18} & VGG-16  &91.3 \\
Low-rank B-CNN \cite{kong2017low}& VGG-16 &90.9 \\
Kernel-Pooling \cite{cui2017kernel}&VGG-16 & 92.4 \\
DFL-CNN \cite{wang2018learning}&VGG-16 &93.3\\
\hline
RA-CNN \cite{fu2017look} & VGG-19  & 92.5\\
MA-CNN \cite{zheng2017learning}& VGG-19  & 92.8 \\
\hline
Kernel-Pooling \cite{cui2017kernel}&ResNet-50 & 91.1\\
MAMC \cite{sun2018multi}& ResNet-50  & 92.8 \\
FT-ResNet \cite{he2016deep}* & ResNet-50  &92.5 \\
DT-RAM \cite{li2017dynamic} & ResNet-50 &93.1\\
DFL-CNN \cite{wang2018learning}& ResNet-50 & 93.1\\
NTS-Net \cite{yang2018learning} & ResNet-50  & 93.9\\
TASN \cite{zheng2019looking} &ResNet-50 & 93.8 \\
DCL \cite{Chen_2019_CVPR} & ResNet-50  & 94.5 \\
Cross-X \cite{luo2019cross} & ResNet-50  &{94.6}\\
LIO \cite{Zhou_2020_CVPR} & ResNet-50  &94.5\\
ACNet \cite{ji2020attention} & ResNet-50  &94.6\\
\hline
 \rowcolor{mygray} {CMN (ours)} & ResNet-50 & {94.9} \\
\hline
\end{tabular}
\end{center}
\label{tab3}
\end{table}
\setlength{\tabcolsep}{8pt}
\begin{table}[t]
\caption{Comparison of various methods in terms of top-1 accuracy (\%) on \emph{non-rigid} \textbf{CUB-200-2011}. * denotes our re-implementation.}
\small
\begin{center}
\begin{tabular}{|l|c|c|c|c|}
\hline
\multicolumn{1}{|l|}{\multirow{2}{*}{Methods}}& \multicolumn{1}{l|}{\multirow{2}{*}{Backbone}}& \multicolumn{1}{l|}{\multirow{2}{*}{Accuracy (\%)}}\\
\multicolumn{1}{|c|}{} & {}  &\\
\hline\hline
BilinearCNN \cite{LinRM18} & VGG-16  &84.1\\
Low-rank B-CNN \cite{kong2017low}& VGG-16 &84.2\\
Kernel-Pooling \cite{cui2017kernel}&VGG-16 &86.2 \\
DFL-CNN \cite{wang2018learning}&VGG-16 &85.8\\
\hline
RA-CNN \cite{fu2017look} & VGG-19 &85.3\\
MA-CNN \cite{zheng2017learning}& VGG-19 &86.5 \\
\hline
FT-ResNet \cite{he2016deep}* & ResNet-50  &85.8 \\
DFL-CNN \cite{wang2018learning}& ResNet-50 &87.4\\
NTS-Net \cite{yang2018learning} & ResNet-50  &87.5\\
TASN \cite{zheng2019looking} &ResNet-50 &{87.9} \\
DCL \cite{Chen_2019_CVPR} & ResNet-50 &87.8 \\
Cross-X \cite{luo2019cross} & ResNet-50  &87.7\\
LIO \cite{Zhou_2020_CVPR} & ResNet-50 &88.0\\
ACNet \cite{ji2020attention} & ResNet-50  &88.1\\
\hline
 \rowcolor{mygray} \rowcolor{mygray}{CMN (ours)} & ResNet-50 & 88.2\\
\hline
\end{tabular}
\end{center}
\label{tab4}
\end{table}

\begin{figure*}[t]
\begin{center}
\includegraphics[width=1\linewidth]{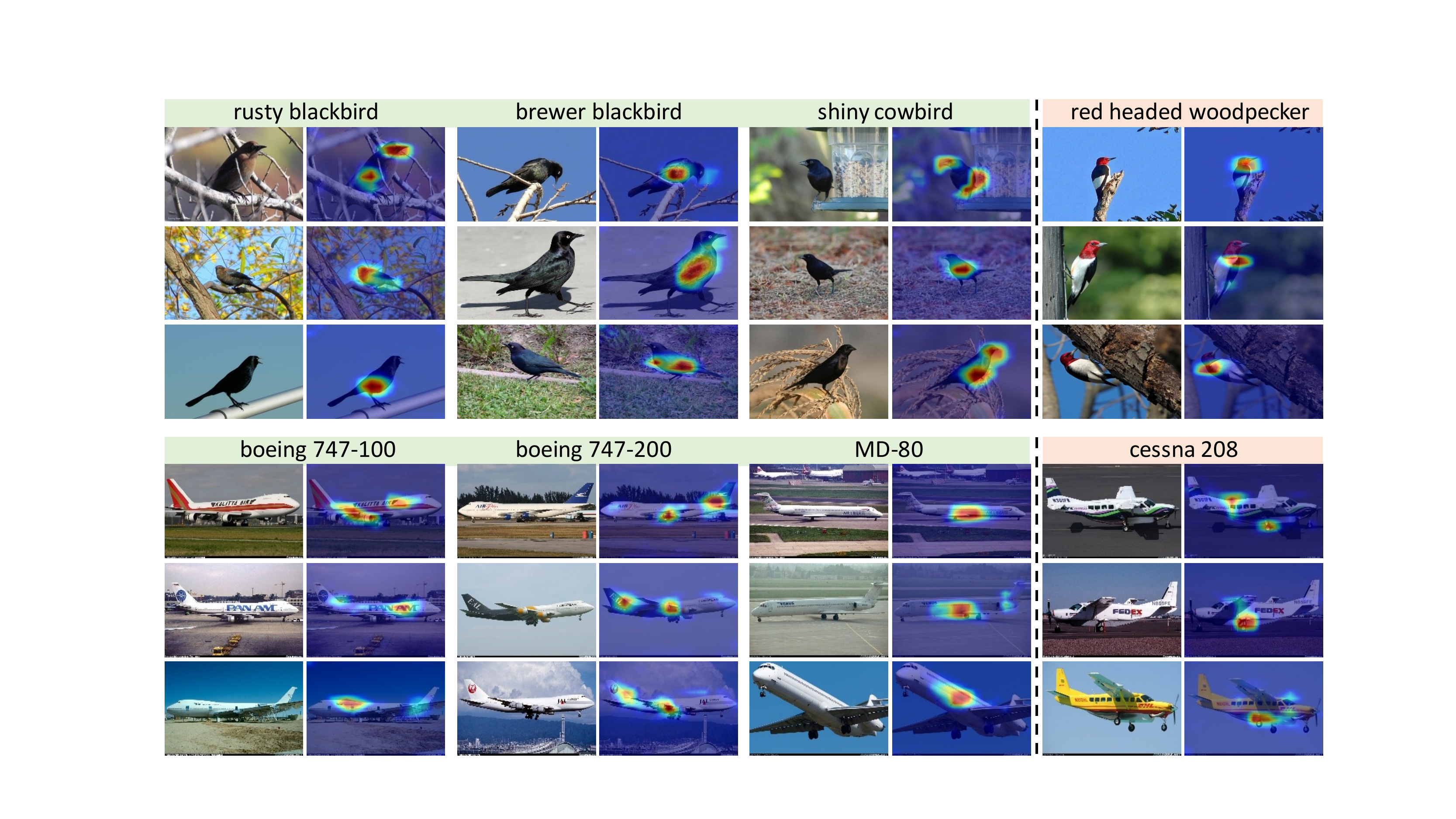}
\end{center}
\caption{Visualization of the learned visual cues on the \emph{rigid} dataset Aircraft and the \emph{non-rigid} dataset CUB. Based on the attention scores computed by our method, we select images from three visually similar classes (the first three columns) and one dissimilar class (the last column) for each dataset. The similar classes share some visual patterns, \emph{e.g.}, ``breast" and ``bill" for birds, and ``under-wing fuselage'', ``tail", and ``upperdeck'' for aircraft.  Each class contains unique and subtle visual cues. Best viewed in color.}
\label{fig_cam}
\end{figure*}
\begin{figure}[t]
    \centering
   {\includegraphics[width=0.95\linewidth]{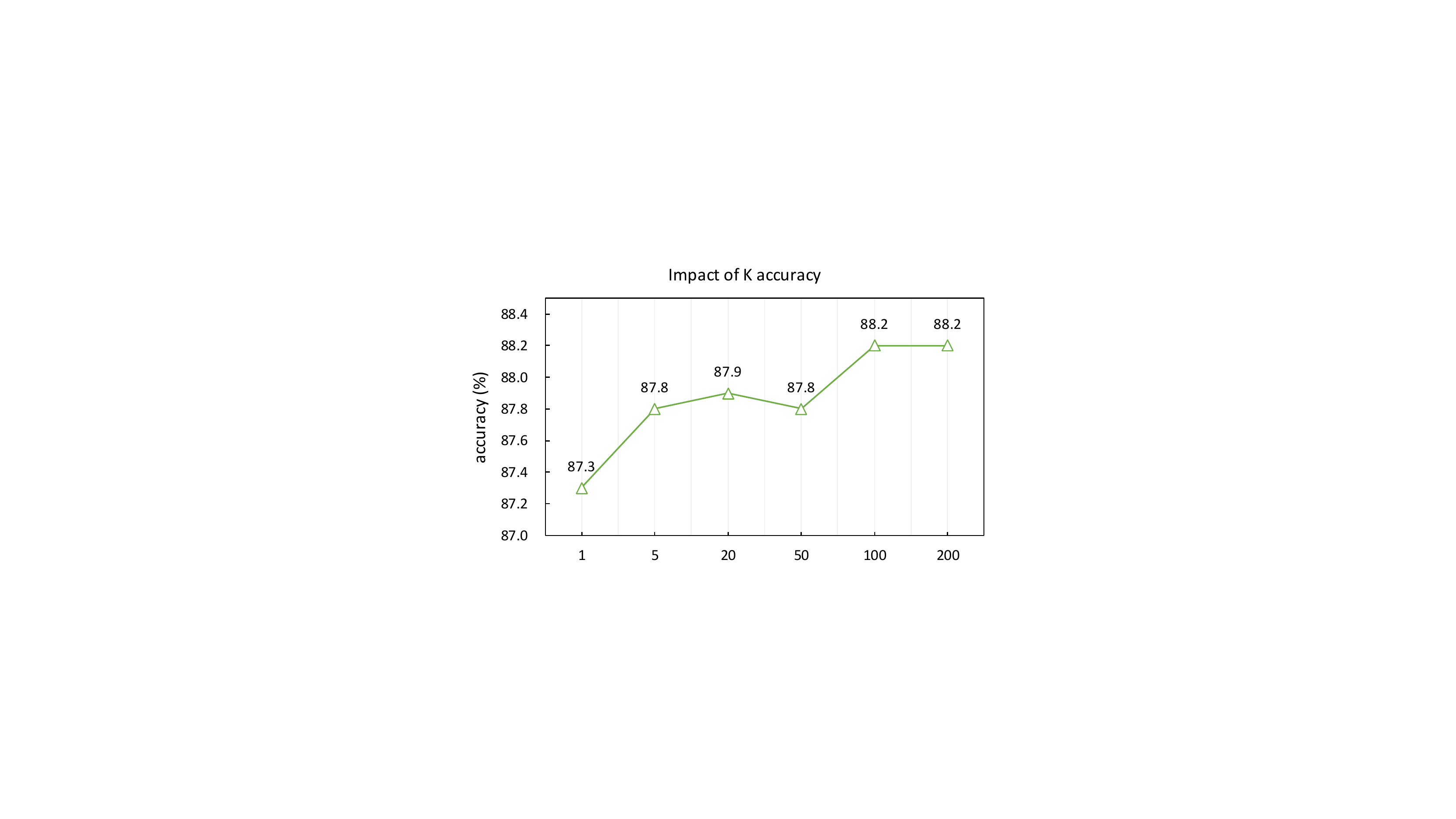}}%
    \caption{Importance of prototype diversity. We show the accuracy (\%) of using only the prototypes with top-K attention scores to form a memory response feature. When K=200, the response is given access to all prototypes. The results are on the CUB-200-2011 dataset.}
    \label{fig_ab}%
\end{figure}

\textbf{Comparison on NABirds.} The architecture of the purposed method is simple and effective. It can easily be applied to large-scale datasets. We also conduct experiments on another non-rigid dataset NABirds and report results in Table \ref{tab5}. In this dataset, we observe that our method also achieves $+3.8\%$ improvement over baseline and arrives at 87.8 \%. This improvement also validate the effectiveness of our method. Moreover, CMN is competitive with the recent state-of-the-art method Cross-X \cite{luo2019cross}. More specific, our method is 1.6\% higher than Cross-X in the same settings.
We think our method considers shared patterns across classes (inter-class similarities) that is a characteristic of the fine-grained classification, and thus achieves improvements over baselines.

\setlength{\tabcolsep}{13pt}
\begin{table}[t]
\caption{Comparison of various methods on \emph{non-rigid} \textbf{NABirds} dataset. Here, * denotes our re-implementation. Our method achieves competitive accuracy with the-state-of-the-art methods.}
\small
\begin{center}
\begin{tabular}{|l|c|c|}
\hline
Method    & Backbone & Accuracy (\%)\\
\hline\hline
BilinearCNN \cite{LinRM18} & VGG-16 & 79.4 \\
FT-ResNet \cite{he2016deep}* & ResNet-50 & 84.0 \\
MaxEnt \cite{dubey2018maximum} & ResNet-50 & 69.2 \\
MaxEnt \cite{dubey2018maximum} & DenseNet-161 & 83.0 \\
Cross-X \cite{luo2019cross} & ResNet-50 & 86.2 \\
\hline
 \rowcolor{mygray}{CMN (ours)}    & ResNet-50    &{87.8}\\
\hline
\end{tabular}
\end{center}
\label{tab5}
\end{table}

\begin{figure}[t]
    \centering
   {\includegraphics[width=0.95\linewidth]{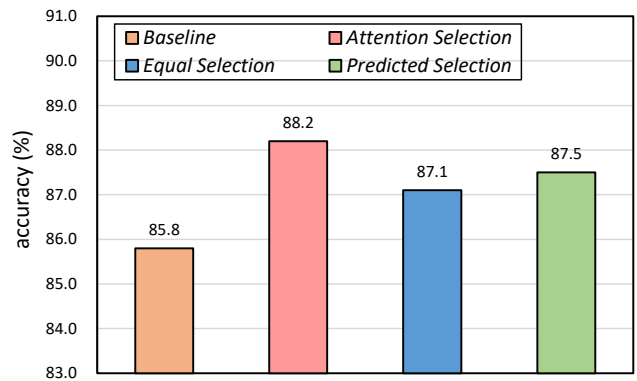}}%
    \caption{Comparison of three attention score mechanisms for memory reading. We test attention-based selection \emph{i.e.} our technique, equal weighting for all prototypes, and predicting scores based on the given image via a fully-connected layer. The results are on the CUB-200-2011 dataset. The proposed attention-based selection is the best way to combine class prototypes.
    }
    \label{fig_ab}%
\end{figure}

\subsection{Component Analysis}
\textbf{Effect of attention-based memory reading.}
Given an input feature, we adopt an attention-based method of retrieving relevant prototypes (as detailed in Section \ref{read}). To validate its effectiveness, we introduce two variants for comparison, 1) equal selection, where all attention scores are equal to $1/C$ ($C$ is the number of categories); 2) predicted selection \cite{li2016learning}, which uses a fully connected layer followed by a softmax to predict attention scores for the given input feature. 
We compare these two variants and our method in Figure \ref{fig_ab}. They both improve the accuracy compared with our baseline. This demonstrates that leveraging class prototypes is beneficial for fine-grained feature learning. Meanwhile, equal selection delivers worse performance than predicted selection (87.1\% vs. 87.5\%), indicating that the retrieved prototypes should be dependent on the input feature.
Our attention-based method outperforms equal section and predicted selection by 1.1\% and 0.7\%, respectively. This is because the attention-based method is based on Euclidean distance, which is a direct and effective way to represent the feature correlation.

\textbf{Importance of learning in the class-specific module.} Class-specific memory module records the moving average of each class and is updated during training. To validate the importance of learning in the module, we test a random module for comparative purposes, \emph{i.e.}, we randomly initialize the memory matrix and fix it during training. In Table \ref{table5}, we compare the two modules on Aircraft and CUB. We observe that the random module is on par with our baseline. This indicates that 1) naively increasing network capacity is not beneficial and 2) that the module should record useful and meaningful information and be updated during network learning.

\textbf{Importance of prototype diversity.} 
We leverage all prototypes to compute a response feature (Eq. \ref{combination}). In Figure \ref{fig_ab}, we show the effect on accuracy of using only the prototypes with the top-K attention scores on CUB. We observe that allowing the response to rely on more than just the most similar prototypes improves accuracy. 
Moreover, we observe that the accuracy does improve using more than 100 prototypes. More specifically, the accuracy does not improve whrn using more than 100 prototypes. This is because the response feature is most formed by the half relevant prototypes, which is consistent with our observation in attention distribution (Figure \ref{fig7}): the most relevant prototypes contribute more information to enhance the input feature.
\setlength{\tabcolsep}{8pt}
\begin{table}[!t]
\caption{Importance of learning in the class-specific module. The class-specific module denotes recording the moving average of each class. The random module means that memory matrix is randomly initialized and fixed during training.}
    \small
    \begin{center}
    \begin{tabular}{|c|c|c|}
    \hline
    Methods & FGVC Aircraft & CUB-200-2011  \\
    \hline
    \hline
    Baseline & 90.0 & 85.8 \\
    \hline
    Random module & 90.0 & 85.8 \\
    \hline
    \rowcolor{mygray} {Class-specific module} & {93.8} & {88.2}\\
    \hline
    \end{tabular}
    \end{center}

 \label{table5}
\end{table}

\setlength{\tabcolsep}{10pt}
\begin{table}[!t]
\caption{Effect of memory module on CIFAR-100. The memory module does not bring any improvement and in fact, seems to hinder performance. This suggests that our method is specifically tailored for fine-grained recognition, where images are semantically and visually similar.}
    \small
    \begin{center}
    \begin{tabular}{|c|c|c|}
    \hline
    Networks & Memory module & Top-1 Accuracy (\%)  \\
    \hline
    \hline
    ResNet-18 &  & 75.42\\
  \rowcolor{mygray} ResNet-18 & \checkmark & 75.12 \\
    \hline
    ResNet-34 &  &  75.54\\
     \rowcolor{mygray}ResNet-34 &\checkmark  & 75.51 \\
    \hline
    ResNet-50 & & 77.10\\
    \rowcolor{mygray}ResNet-50 & \checkmark&76.70 \\
    \hline
    \end{tabular}
    \end{center}
    \label{cifar-100}
\end{table}

 \begin{figure}[h]
 \begin{center}
\includegraphics[width=1\linewidth]{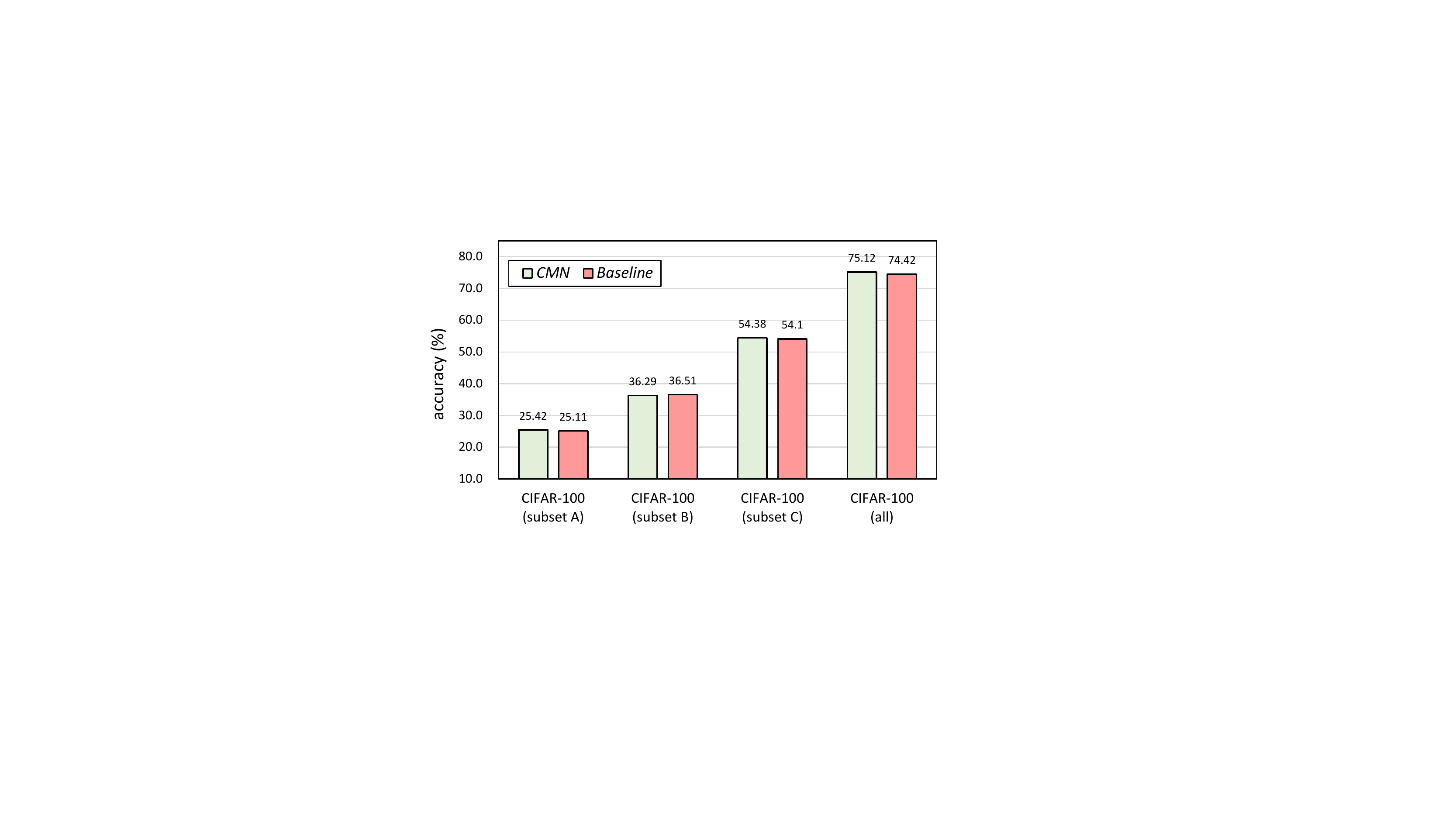}
\caption{Comparison between ResNet-18 and CMN under different training sets. CIFAR-100 contains all 50,000 training images; CIFAR-100 (subset A) has 5,000 training images, CIFAR-100 (subset B) has 10,000 training images, CIFAR-100 (subset C) has 25,000 training images. We find that both methods are on par with each other under four training sets. This indicates CMN is not suitable for general datasets where the inter-class similarity is relatively small.}
\label{fig8}
\end{center}
\end{figure}


%
\textbf{Feature visualization.}  We average the attention scores for the images of each class. For a particular class, a higher average attention score indicates that the corresponding class is semantically similar. 
In Figure \ref{fig_cam}, we visualize the feature maps of last convolutional layer. We select the three most relevant classes (the first three columns) as defined by our method and one irrelevant class (the last column). We observe that images of correlated classes tend to focus of similar patterns, \emph{e.g.}, \emph{Shiny Cowbird}, \emph{Brewer Blackbird}, and \emph{Rusty Blackbird} have the same visual evidence of breast pattern. This shared pattern can effectively distinguish them from irrelevant \emph{Red-headed Woodpecker}. 
Moreover, each class has it own distinctive cues, \emph{e.g.}, \emph{Boeing 747-200} has a discriminative evidence on ``tail" part, and \emph{Cessna 208} has an unique pattern on ``wheel” part.

\subsection{Further Evaluation.} We also conduct experiments on more general classification dataset CIFAR-100 \cite{krizhevsky2009learning}, where the inter-class similarity is relatively small. In Table \ref{cifar-100}, we validate the effect of the memory module on several networks. We observe that our CMN does not bring any improvement on this dataset. This is because the classes in CIFAR-100 have low class-class similarity.
Furthermore, our method could add some unnecessary class-class cues to the input feature, the learning difficulty may be increased, thus having slightly lower accuracy.

In fine-grained classification datasets, the number of training images per class is relatively small. In comparison, CIFAR-100 contains adequate training data for each class. In addition, the class-specific prototypes have shown its effectiveness for few-shot learning \cite{hsu2018unsupervised,snell2017prototypical,wertheimer2019few}.
This motivates us to further study the working mechanism of CMN under the few-example setting. Specifically, we study whether CMN mainly helps when training data per class is scarce on CIFAR-100.

To validate this, we reduce the number of training images in CIFAR-100 and report the experimental results in Figure \ref{fig8}. In practice, we create three subsets, 1) CIFAR-100 (subset A) has 5,000 training images; 2) CIFAR-100 (subset B) has 10,000 training images; 3) CIFAR-100 (subset C) has 25,000 training images. 
As shown in Figure \ref{fig8}, we observe that both CMN and original ResNet-18 achieve almost the same top-1 accuracy under the different CIFAR-100 subsets. This also indicates that CMN might not suitable for general classification dataset that contains relatively small inter-class similarity.
Thus, we believe the advantage of CMN lies in that it has ability to leverage the shared semantic information across classes to well represent the input image. 
In summary, CMN is specifically tailored for fine-grained classification, where images are both semantically and visually similar.

\section{Conclusions}
We introduce an efficient, effective, and interpretable class-specific memory module for fine-grained visual classification. By storing the prototypical information of all classes, the memory module allows the network to share semantic information between classes. We use an attention-based method to retrieve relevant class prototypes for a given input and generate a specially tailored response feature to augment the original feature. We integrate the memory module with a convolutional neural network, yielding a Categorical Memory Network (CMN). Our method leverages relevant prototypes to produce representations that express inter-class semantics while preserving instance-specific discriminative cues. Moreover, encouraging the network to represent inter-class similarities leads to more robust fine-grained representations. 

Our method significantly increases accuracy at effectively no additional computational cost, achieving accuracy competitive with state-of-the-art methods on four fine-grained classification benchmarks. In the future, we plan to explore unsupervised categorical memory networks where the aggregation of prototypes may allow for the discovery of unsupervised sub-categories.

\bibliographystyle{IEEEtran}
\bibliography{IEEEabrv,REFS}
\end{document}